\documentclass[conference]{IEEEtran}
\IEEEoverridecommandlockouts
\usepackage{cite}
\usepackage{graphicx}
\usepackage[cmex10]{amsmath}
\usepackage{amssymb}

\usepackage{url}
\usepackage{tikz}

\usepackage{tabularx}
\usetikzlibrary{bayesnet}
\usetikzlibrary{arrows,positioning}
\usepackage{siunitx}
\usepackage{booktabs}
\usepackage{comment}
\usepackage{changepage}
\usepackage{float}
\usepackage{subfig}
\usepackage{hyperref}

\newcommand{\etal}{\emph{et al.~}}

\begin{document}
\title{A Factor Graph Approach to Multi-Camera Extrinsic Calibration on Legged
Robots}


%


\author{\IEEEauthorblockN{Andrzej
Reinke$^{*,\dagger}$}\IEEEauthorblockA{\href{mailto:andreink@student.pg.gda.pl}{
andreink@student.pg.gda.pl}}\and
\IEEEauthorblockN{Marco Camurri$^{*,\ddagger}$}
\IEEEauthorblockA{\href{mailto:mcamurri@robots.ox.ac.uk}{
mcamurri@robots.ox.ac.uk}} \and
\IEEEauthorblockN{Claudio Semini$^{*}$}
\IEEEauthorblockA{\href{mailto:claudio.semini@iit.it}{claudio.semini@iit.it}}
\thanks{\noindent This work was supported by Istituto Italiano di Tecnologia
(IIT)}
\IEEEcompsocitemizethanks{\IEEEcompsocthanksitem $^{*}$ Dynamic
Legged Systems Lab, Istituto Italiano di Tecnologia, Italy.
\IEEEcompsocthanksitem$^{\dagger}$ Gda\'nsk University of
Technology,
Poland.
 \IEEEcompsocthanksitem$^{\ddagger}$ Oxford Robotics Institute,
 University of Oxford, UK.
}}

\maketitle

\begin{abstract}
Legged robots are becoming popular not only in research, but
also in industry, where they can demonstrate their superiority over
wheeled machines in a variety of applications.
Either when acting as mobile manipulators or just as all-terrain ground
vehicles, these machines need to precisely track the desired base and
end-effector trajectories, perform Simultaneous Localization and Mapping
(SLAM), and move in challenging environments, all while keeping
balance. A
crucial aspect for these tasks is that all onboard sensors must be
properly calibrated and synchronized to provide consistent
signals for all the software modules they feed.
In this paper, we focus on the problem of calibrating the relative pose
between a set of cameras and the base link of a quadruped robot. This
pose is fundamental to successfully perform sensor fusion, state estimation,
mapping, and any other task requiring visual feedback.
To solve this problem, we propose an approach based on factor graphs
that jointly optimizes the mutual position of the cameras and the robot
base using kinematics and fiducial markers.
We also quantitatively compare its performance with other state-of-the-art
methods on the hydraulic quadruped robot HyQ. The proposed approach is simple,
modular, and independent from external devices other than the fiducial marker.
\end{abstract}


%
\IEEEpeerreviewmaketitle

\section{Introduction}

Robot calibration is a widely explored subject that involves the
estimation of kinematics and camera intrinsic/extrinsic parameters, either by
optical
or mechanical methods.

\emph{Kinematics calibration} adjusts robot parameters such as: link lengths,
angle
offsets and other body dimensions that are crucial for precise end-effector
placement, especially in industry \cite{BookIndustrial}.
These adjustments are required due to imperfections in
manufacturing of mechanical parts, their assemblies, and aging processes.

\emph{Intrinsic calibration} retrieves the parameters of the so-called camera
pinhole model, which projects 3D points in space onto the image plane of the
camera. These parameters model the skewness of the CMOS elements, lens placement
and the optical center offsets. Since the lenses have finite dimension and non-null thickness, the model is typically extended with additional parameters to
compensate for tangential and radial distortions as well.

Finally, \emph{extrinsic calibration} consists on finding the pose (i.e.,
translation
and rotation) of the camera with respect to a specific object captured within
its field of view. In robotics, it is also important to estimate
the camera pose expressed in the robot base frame. We will refer
to
this process as \emph{robot-camera extrinsic} calibration\footnote{This is
similar to the so-called ``hand-eye calibration'', where the relative pose
between a camera mounted on an arm and a robotic hand has to be
detected\cite{horaud95ijrr}. In our case, there is no actuated linkage between
the camera and  the base, even though leg kinematics is used to perform
the calibration.}. This tells the robot
where its ``eyes'' are, enabling it to track external objects and achieve  true
spatial cognition. An estimate of this robot-camera extrinsic directly
influences all tasks involving sensor fusion of proprioceptive and exteroceptive
signals, including state estimation, mapping, planning, and control.

Unfortunately, estimating robot-camera extrinsic is not an easy task. In
prototyping applications, sensors are often replaced with new models due to
obsolescence, hence
their mounting locations are not always known by CAD design. On the other hand,
mature products will still suffer from wear and tear or manufacturing defects.
Therefore, it is desirable to have an easy, fast and accurate method to
automatically perform this type of calibration whenever possible.

In this paper, we propose a factor graph based approach to robot-camera
calibration for multiple cameras mounted on a legged robot. Factor graphs have
been successfully applied to several inference problems \cite{factor}, such as:
SLAM, 3D reconstruction, and spatiotemporal crop monitoring. This work shows how
they are suitable to perform robot-camera extrinsic calibration as well.

The contributions of the paper can be summarized as follows:
\begin{itemize}
\item We present a novel approach to multiple robot-camera calibration
based on factor graph optimization; the factors are designed as constraints
originated by kinematics and visual detections of fiducial markers  mounted at
the robot's end-effector. The approach is modular and adaptable to any
articulated
robot with an arbitrary number of cameras and limbs;
\item We provide a quantitative comparison between our method and other
state-of-the-art methods available to date. To the best of our knowledge, this
is the first time such comparison is presented to the scientific community;
\item We deliver the implementation of this method as an \mbox{open-source} ROS
package\footnote{See
\url{https://github.com/iit-DLSLab/dls_factor_graph_calibration}}.
\end{itemize}

The paper outline is the following: Section \ref{ch:related_work} presents the
related works on extrinsic camera calibration for mobile robots; Section
\ref{ch:theory} formally describe the problem of robot-camera extrinsic
calibration; Section \ref{sec:multi-camera-factor} describes our factor
graph approach
for extrinsic and multi-camera calibration; Section \ref{ch:experiments}
states
the result of this work and compares our methods with the
state-of-the-art; Section \ref{ch:conclusion_future_work} summarizes the
work, and points to future directions.

\section{Related work}
\label{ch:related_work}
Extrinsic calibration in robotics has been investigated since the
introduction of visual sensors to industrial manipulators. In this section, we limit to the literature dedicated to mobile robots, which can be broadly
divided into  visual-kinematics  methods and visual-inertial methods.

\subsection{Visual-Kinematics Methods}
The most common extrinsic camera calibration technique for mobile robots
use an external marker (i.e. a checkerboard) placed at the end-effector. The
procedure demands to move the robot arm constantly, recognize marker key points,
and estimate the relative transformations to the camera. The collected
transformations allow to compare camera marker poses with the poses estimated
through the kinematic chain, according to encoder values. Then, the cost
function to be minimized is formulated as the reprojection error between the
detected marker and its projection through the kinematic chain.

Pradeep \etal \cite{pradeep14exprob} proposed an approach based on bundle
adjustment to calibrate the
kinematics, lasers and camera placements on the PR2 robot. The optimization
includes all transformations needed for the robot (sensors to sensors and
sensors to end-effectors) and the measurement uncertainties are modeled as
Gaussians. The major drawback of the procedure is the calibration
time, which is 20 minutes onboard the PR2, plus 25 minutes for the offline
nonlinear optimization. This largely limits the application
during field operations.

More recently, Bl\"ochlinger \etal \cite{bloechlinger16clawar} applied the same
idea to the quadruped robot StarlETH. They used a couple of circular fiducial
marker arrays attached to the front feet and minimized the reprojection error
between the detected markers (with a monocular camera) and their projection
using forward kinematics. The objective of the cost function  includes a set of
33 parameters, including: link lengths, encoder offsets, time offsets, and also
the robot-camera relative pose. Since the focus was more on finding the
kinematics parameters, no results are shown for the robot-camera calibration.
The performance on link lengths is assessed by comparison with the CAD
specifications, yielding a margin of \SI{5}{\milli\meter}.

To reduce the uncertainty due to kinematics, in our previous work
\cite{camurri15mfi}, we developed a method based on fiducial markers and an
external capture system. The fiducial marker is not attached to the robot body,
but it is placed externally, surrounded by a circular array of Motion Capture
(MoCap) markers (small
reflective spheres). The MoCap markers are also placed on the robot at a
known location. When the camera detects the fiducial marker, the robot-camera
 affine transform is recovered from the chain formed by the
camera-to-fiducial,
the vicon-to-fiducial and the robot-to-vicon transforms. An obvious drawback of
this method is the need of an expensive equipment for MoCap.

\subsection{Visual-Inertial Methods}
In the visual-inertial calibration domain, the objective is the relative
transformation between an IMU and a camera. If the relative pose between IMU and
base link is known, this method is equivalent to robot-camera calibration. Since
no kinematics is involved, the primary application is Micro-Aerial-Vehicles
(MAVs): in this context, the different frequencies and time delays between
inertial and camera data need to be taken into account \cite{kelly14tro}.

Lobo and Dias \cite{lobo07ijrr} described a two-step, IMU-camera calibration
procedure. First, they calculate the relative orientation by detecting the
gravity vector with the IMU and the lines of a vertically placed marker with the
camera. Then, the translation is calculated as the lever arm separating the two
sensors, which are vertically aligned from the previous step and moved on a
turntable. The main disadvantage of this method is its dependency on controlled
setups, which are unavailable for sensors rigidly mounted on a mobile robot.
Moreover, ignoring the mutual correlation between rotation and translation can
lead to further errors.

Mirzaei and Rumeliotis \cite{mirzaei08tro} proposed a filter-based algorithm
to overcome some limitations of \cite{lobo07ijrr}. The method is also two-step.
First, they create a prior for the IMU-camera transform by using observations
of an external marker for the camera pose and from manual measurement for the IMU
pose, respectively. Second, they refine the initial guess by performing
visual-inertial state estimation with an Extended Kalman Filter (EKF). The
state includes: position and linear velocity of the IMU, the IMU biases, and
the unknown transformation between the IMU frame and the camera frame. The
authors also demonstrate that these states are observable.

Similarly to \cite{mirzaei08tro}, Kelly \etal \cite{kelly09cira} proposed an
approach based on the Unscented Kalman Filter (UKF), which has been tested
both with and without prior knowledge of a calibration marker.

Furgale \etal \cite{furgale13iros} proposed an estimator for IMU-camera
calibration that simultaneously determines the transformation and the temporal
offset between the two sensors. In contrast to previous methods, the
time-varying states are represented as the weighted sum of B-spline functions,
so that the state can be considered continuous. Their proposed implementation,
\emph{Kalibr}, is publicly available and it has been included in our
comparative study.

Visual-Inertial extrinsic calibration is independent from robot kinematics.
However, it requires homogeneous and smooth excitation on all the IMU axes to
perform well. Even though this is simple for small quadrotors (manually moved in
the air), it becomes not trivial for large legged robots.
Furthermore, the precise location of the IMU on the robot is assumed as known.

\section{Problem Statement}
\label{ch:theory}
\begin{figure}
	\centering

	\subfloat[][\emph{Frames of Reference}]
{\includegraphics[height=4.5cm]{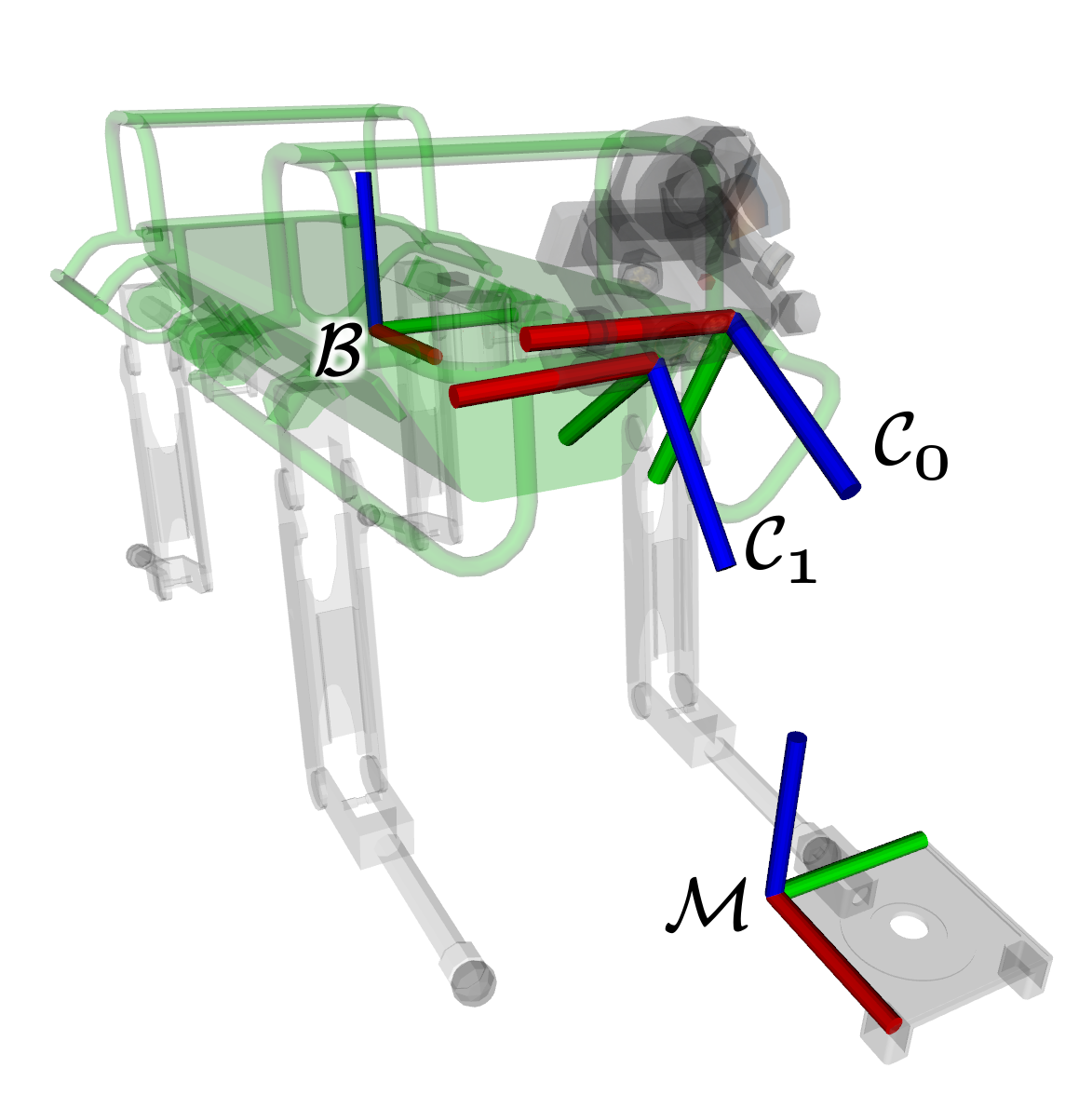}%
\label{fig:ReferenceFrames}} \quad
\subfloat[][\emph{Calibration Marker}]
{\includegraphics[height=4.5cm]{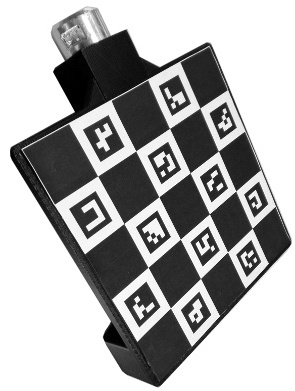}\label{%
fig:aruco-marker}}
\caption{(a) Reference frames for HyQ extrinsic calibration.
$\mathcal{B}$ is the
robot base link; $\mathcal{M}$ is the fiducial marker reference
frame; $\mathcal{C}_{0}$, $\mathcal{C}_{1}$ represent the optical center of  the
Multisense SL left camera and the RGB-D ASUS, respectively. (b) The
special foot with the ChAruco marker \cite{romero18ivc}.
    }
\end{figure}

Let us consider a floating base, articulated robot with a main body, one or 
more limbs, and one or more cameras rigidly attached to the main body. Our goal
is to retrieve the relative pose between the optical center $\mathcal{C}_i$ of
the cameras and the base frame $\mathcal{B}$ on to the main body, by means
of: a) kinematic constraints and b) the detection of a fiducial marker located
at the end-effector of a limb, with frame $\mathcal{M}$ (see Fig.
\ref{fig:ReferenceFrames}).

\begin{figure}
	\centering

\includegraphics[height=4.5cm]{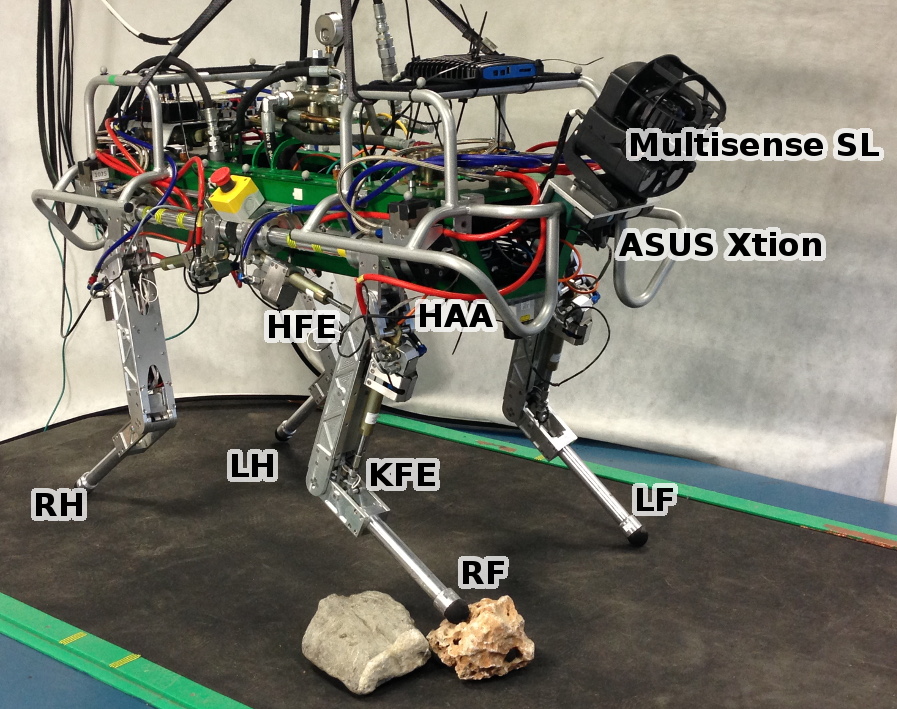} %
\includegraphics[height=4.5cm]{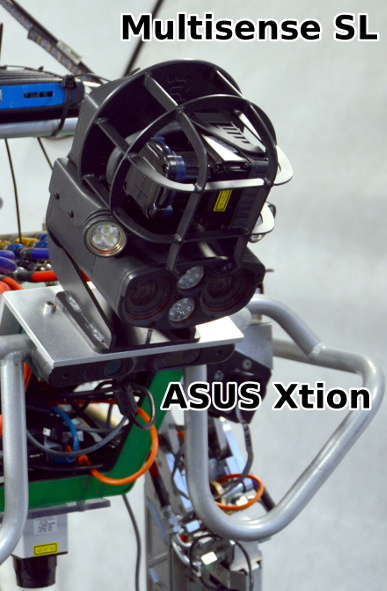}
	\caption{HyQ naming conventions and sensors. Each leg has three DoFs:
Hip Abduction-Adduction (HAA), Hip Flexion-Extension (HFE),
and the Knee Flexion-Extension (KFE). The legs are named as follows: Left Front
(LF), Right Front (RF), Left Hind (LH),
Right
Hind (RH). The robot is equipped with a Multisense SL stereo (and LiDAR, not
used) sensor and an ASUS Xtion RGB-D sensor.}
	\label{fig:robotHyq}
\end{figure}

In this paper, we will consider the HyQ robot \cite{DesignHyq} as the
experimental platform of choice. HyQ is a \SI{90}{\kilo\gram},
\SI{1}{\meter} long hydraulic quadruped robot with 12
actuated Degrees-of-Freedom (DoFs). Each leg has 3 DoFs (Fig.
\ref{fig:robotHyq}): the Hip
Abduction-Adduction (HAA), the Hip Flexion-Extension (HFE) and the Knee
Flexion-Extension (KFE). The reference frames are
shown in Fig.
\ref{fig:ReferenceFrames}: the base frame $\mathcal{B}$ is conventionally placed
at the torso center, with the $xy$-plane passing through the HAA motor axes;
the frames $\mathcal{C}_0$ and
$\mathcal{C}_1$ are the optical frames of the Multisense SL's left camera and the ASUS
Xtion's RGB camera, respectively; the marker frame $\mathcal{M}$ is at the
bottom right corner of the ChAruco
fiducial marker \cite{romero18ivc} mounted on the tip of HyQ's LF foot (Fig.
\ref{fig:aruco-marker}).

\section{Multi-Camera Factor Graph Calibration}
\label{sec:multi-camera-factor}

\begin{figure}
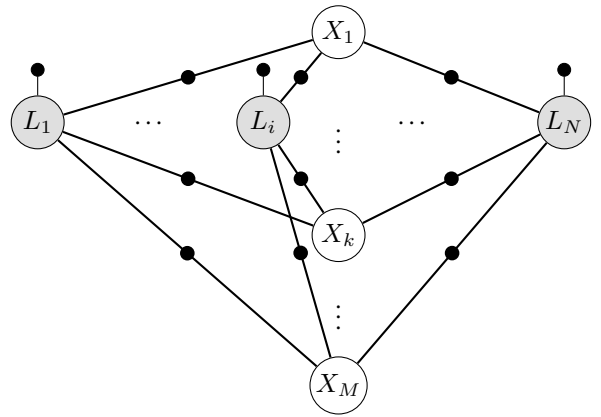

\begin{center}
\tikz{ %
\node[latent, xshift=1.0cm,yshift=-1.3cm] (X_1) {$X_1$} ; %
\node[latent, xshift=1.0cm,yshift=-4.0cm] (X_k) {$X_k$} ; %
\node[latent, xshift=1.0cm,yshift=-6.0cm] (X_M) {$X_M$} ; %
\node[obs,xshift=0cm, yshift=-2.5cm] (L_i) {$L_i$} ; %
\node[obs, xshift=4cm,yshift=-2.5cm] (L_N) {$L_N$} ; %
\node[obs, xshift=-3cm,yshift=-2.5cm] (L_1) {$L_1$} ; %
\node[midway,circle,draw,fill=black,xshift=-3cm,yshift=-1.8cm,scale=0.1] (factor_L1)
{$factor_L1$};
\node[midway,circle,draw,fill=black,xshift=0cm,yshift=-1.8cm,scale=0.1] (factor_L2)
{$factor_L1$};
\node[midway,circle,draw,fill=black,xshift=4cm,yshift=-1.8cm,scale=0.1] (factor_L3)
{$factor_L1$};
\path (X_1) -- node[auto=false]{\vdots} (X_k);
\path (X_k) -- node[auto=false]{\vdots} (X_M);
\path (L_i) -- node[auto=false]{\ldots} (L_N);
\path (L_1) -- node[auto=false]{\ldots} (L_i);
\draw[-] (L_1) -- (factor_L1);
\draw[-] (L_i) -- (factor_L2);
\draw[-] (L_N) -- (factor_L3);
\draw [thick, black!10!black] (X_1) -- +(L_i) node[midway,circle,draw,fill=black,scale=0.5] {};
\draw [thick, black!10!black] (X_k) -- +(L_i) node[midway,circle,draw,fill=black,scale=0.5] {};
\draw [thick, black!10!black] (X_1) -- +(L_N) node[midway,circle,draw,fill=black,scale=0.5] {};
\draw [thick, black!10!black] (X_k) -- +(L_N) node[midway,circle,draw,fill=black,scale=0.5] {};
\draw [thick, black!10!black] (X_M) -- +(L_1) node[midway,circle,draw,fill=black,scale=0.5] {};
\draw [thick, black!10!black] (X_M) -- +(L_i) node[midway,circle,draw,fill=black,scale=0.5] {};
\draw [thick, black!10!black] (X_M) -- +(L_N) node[midway,circle,draw,fill=black,scale=0.5] {};
\draw [thick, black!10!black] (L_1) -- +(X_1) node[midway,circle,draw,fill=black,scale=0.5] {};
\draw [thick, black!10!black] (L_1) -- +(X_k)
node[midway,circle,draw,fill=black,scale=0.5] {};
}
\caption{The Factor Graph representing the calibration of $M$ cameras over $N$
image poses. The gray ``landmark'' nodes $L_i$ represent the poses of the
fiducial marker acquired by the cameras. The unary factors from these nodes
correspond to the forward kinematics constraint to the base,  computed
from the encoder values. The nodes $X_i$ represent the unknown
poses
of the camera in
the base frame. They are joined to the landmark nodes by the factors computed
from the visual detection of the fiducial markers.}
\label{fig:more-camera-calibratio}
\end{center}
\end{figure}

The factor graph for multi-camera calibration is shown in
Fig. \ref{fig:more-camera-calibratio}. In its general formulation,
an array of $M$ cameras is mounted on the robot. Their pose from the base
frame $\mathcal{B}$ are expressed by the nodes $X_i,\; i\in \mathbb{N}_{M}$.

The gray nodes $L_i,\; i \in \mathbb{N}_{N}$ denote the $N$ observed fiducial
marker poses visible from the cameras over time (the cameras are not required to
detect all the observations). We introduce two types of factors: the
$\emph{kinematics}$ factors and the $\emph{fiducial}$ factors. The
former are unary factors associated to the landmarks and express the kinematic
constraints between the marker frame $\mathcal{M}$ and the base frame
$\mathcal{B}$ by means of forward kinematics. The latter connect $X_i$ with
$L_i$ and correspond to the
relative pose between the fiducial marker and the camera.

For all measurements, a zero-mean Gaussian noise model $N(0,\Sigma)$ is
adopted, where the covariance $\Sigma \in \Lambda^6$ is a diagonal matrix whose
values are computed taking to account: a) the encoder uncertainty and their
propagation to the end effector \cite{hartley18icra}; and b) the uncertainty on
the vertex jitter of the marker detection \cite{romero18ivc}.
After accumulating $N$ marker poses, a Gauss-Newton factor graph optimization is
performed.

\section{Experimental Validation}
\label{ch:experiments}
We have implemented the graph from Section
\ref{sec:multi-camera-factor} on the HyQ robot, using the GTSAM library
\cite{factor}. Given the flexibility of the approach, new cameras or markers
can be added just by including the corresponding nodes in the graph. The
kinematics factors are given by the forward kinematics through the RobCoGen
library \cite{frigerio16joser}. The fiducial factors are implemented with the
AruCo library \cite{romero18ivc}, which is capable of sub-pixel accuracy and
provides pose estimation markers (ChAruco).

In this section, we compare the performance of this approach with the
MoCap-based method of \cite{camurri15mfi} and Kalibr
\cite{furgale13iros}.

The ground truth was collected by physically constraining
the robot to a known configuration and detecting a
\SI{600}{\centi\meter\squared} marker with a known pose (accurately measured
with digital calibers and inclinometers) from the robot.

Table \ref{tab:one} shows the linear and angular Mean Absolute Error (MAE)
obtained for the two cameras mounted on
HyQ.

For Kalibr, we have collected a dataset of 2778 frames
(\SI{526}{\second}), during which the
robot was excited (either manually or with some actuation) on all axes. In
general, the mean results are accurate (linear and angular errors for the
Multisense SL are \SI{2.3}{\percent} and \SI{2.7}{\percent}, respectively),
even though the effort to excite the
robot properly makes the procedure difficult to repeat (hence the high standard
deviation). We believe the low resolution of the ASUS is the major cause of
the poorer results. With the kinematics calibration, the marker can get
much closer to the camera, reducing this effect.

For the MoCap method, we collected 562 frames (\SI{142}{\second}) of the
static robot facing the marker. For the Multisense, the method shows a
\SI{3.4}{\percent} error in both
translation and rotation. Since the MoCap has
sub-millimeter
accuracy, the marker placement (either on the robot
and on the fiducial marker) is the major source of error.

To test the effect of multiple camera constraints, we tested the factor graph
method both for individual cameras and with both cameras at the same time. We
collected a dataset of 1312 images for a total duration of \SI{348}{\second}
while the leg was passively moved in front of both cameras. In general, both
method show slightly worse but more stable and balanced results than Kalibr.
In particular, the mutual
constraint between the cameras reduced the error for the Multisense setup
from \SI{3.9}{\percent} to \SI{2.03}{\percent} on translation
and from \SI{5.7}{\percent} to \SI{3.8}{\percent} in rotation (cf.
F1 and F2 in Table \ref{tab:one}).

\begin{table}
\resizebox{\columnwidth}{!}{%
\centering
\begin{tabular}{lcccccc}

\toprule
\multicolumn{7}{c}{\textbf{Multisense SL}} \\
\toprule
 &\textbf{X ($\boldsymbol{\sigma}$)} & \textbf{Y ($\boldsymbol{\sigma}$)}
&\textbf{Z ($\boldsymbol{\sigma}$)} & \textbf{Roll ($\boldsymbol{\sigma}$)} &
\textbf{Pitch ($\boldsymbol{\sigma}$)} &
\textbf{Yaw ($\boldsymbol{\sigma}$)} \\
&[\si{\centi\meter}]&[\si{\centi\meter}]&[\si{\centi\meter}]&[deg]&[deg]&[deg]\\
\midrule
\textbf{K} & 0.2 (1.8) & 0.2 (1.4)& 1.8 (0.5) & 0.1 (0.3)& 0.8 (0.7) & 0.4 (0.6)\\
\midrule
\textbf{M} & 1.3 & 1.4 & 1.8 & 1.0 & 0.3 & 0.4\\
\midrule

\textbf{F1} & 1.1 (0.1) & 0.9 (0.5)& 2.1 (0.5)& 0.5 (0.2)& 0.2 (0.1)& 1.8 (0.5)\\
\midrule
\textbf{F2} & 1.2 (0.5)  & 0.3 (0.5) & 1.0 (0.2)   & 0.2 (0.5)  & 0.6 (0.2)  & 1.1 (0.2) \\
\toprule
\multicolumn{7}{c}{\textbf{ASUS Xtion}} \\
\toprule
\textbf{K} & 1.2 (1.8) & 3.1 (0.4) & 5.3 (0.5)& 2.9 (0.3)&
3.8 (0.7) & 2.5 (0.6)\\
\midrule
\textbf{F1} & 2.5 (0.3) & 0 (0.4) & 0.1 (0.2) & 0.9 (1.4) &
1.3 (0.2)
& 2.1 (1.2) \\
\midrule
\textbf{F2} & 1.8 (0.4) & 0.6 (0.5) & 0.4 (0.4) & 0.8 (0.5) & 0.2 (0.1)
& 0.2 (0.2)\\
\bottomrule
\end{tabular}}
\caption{Experimental comparison between different
calibration methods. We evaluate the Mean Absolute Error and its standard
deviation on the Multisense SL (top) and ASUS Xtion (bottom):
Kalibr
\cite{furgale13iros} (K), MoCap \cite{camurri15mfi} (M), Factor Graph with
single camera (F1) and Factor Graph with two cameras (F2). For the ASUS, the
MoCap method was not available due to the limited Field of View (FoV) of the
sensor.}
\label{tab:one}
\end{table}



\section{Conclusions and Future Work}
\label{ch:conclusion_future_work}
In this paper, we have presented a factor graph approach to multi-camera
calibration of a quadruped robot. We have demonstrated that a factor graph
framework represents a valid and flexible alternative to visual-inertial methods,
which require smooth motion and balanced excitation of all axes to provide
reliable results. On the other hand, visual-kinematics methods require hardware
modifications at the end-effector (i.e., a fiducial marker support) to be
precise.

Future developments from this work are oriented towards
the integration of both methods in a factor graph fashion to
make the calibration even more robust and automated.

\bibliographystyle{IEEEtran}

\bibliography{areinke19irc}
\end{document}